\documentclass[10pt,conference]{IEEEtran}
\IEEEoverridecommandlockouts
\usepackage{cite}
\usepackage{amsmath,amssymb,amsfonts}
\usepackage{amsmath,amsfonts}
\usepackage{algorithmic}
\usepackage{graphicx}
\usepackage{textcomp}
\usepackage{rotating}
\usepackage{array}
\usepackage{xcolor}
\usepackage{tikz}
\usepackage{makecell}
\usepackage{multirow}
\usepackage[T1]{fontenc}
\usepackage{comment}
\usepackage{etoolbox}
\usepackage{enumitem}
\def\BibTeX{{\rm B\kern-.05em{\sc i\kern-.025em b}\kern-.08em
    T\kern-.1667em\lower.7ex\hbox{E}\kern-.125emX}}

\begin{document}

\newcommand*\circled[1]{\tikz[baseline=(char.base)]{
		\node[shape=circle,draw,inner sep=0.8pt] (char) {#1};}}
	

\title{Statement-based Memory for Neural Source Code Summarization}

\author{\IEEEauthorblockN{Aakash Bansal, Siyuan Jiang, Sakib Haque, and Collin McMillan}
\IEEEauthorblockA{\textit{Dept. of Computer Science and Engineering} \\
	\textit{University of Notre Dame}\\
	Notre Dame, IN, USA \\
	\{abansal1, sjiang1, shaque, cmc\}@nd.edu
}}



\maketitle

\begin{abstract}
Source code summarization is the task of writing natural language descriptions of source code behavior.  Code summarization underpins software documentation for programmers.  Short descriptions of code help programmers understand the program quickly without having to read the code itself.  Lately, neural source code summarization has emerged as the frontier of research into automated code summarization techniques.  By far the most popular targets for summarization are program subroutines.  The idea, in a nutshell, is to train an encoder-decoder neural architecture using large sets of examples of subroutines extracted from code repositories.  The encoder represents the code and the decoder represents the summary.  However, most current approaches attempt to treat the subroutine as a single unit.  For example, by taking the entire subroutine as input to a Transformer or RNN-based encoder.  But code behavior tends to depend on the flow from statement to statement.  Normally dynamic analysis may shed light on this flow, but dynamic analysis on hundreds of thousands of examples in large datasets is not practical.  In this paper, we present a statement-based memory encoder that learns the important elements of flow during training, leading to a statement-based subroutine representation without the need for dynamic analysis.  We implement our encoder for code summarization and demonstrate a significant improvement over the state-of-the-art.
\end{abstract}


\section{Introduction}

The lynchpin to a large portion of software documentation for programmers is the ``source code summary.''  A summary is a natural language description of the behavior of a section of source code.  Program subroutines are the typical target for summaries.  Documentation engines such as JavaDocs and Pydocs use these summaries to help programmers gain a quick understanding of what each subroutine does~\cite{Kramer:1999:ADS:318372.318577}.  Software engineering researchers have long dreamt of automating code summarization, because writing summaries by hand is expensive, especially for legacy systems which source code is often not well-documented~\cite{allamanis2018survey, forward2002relevance, haiduc2010supporting}.

The workhorse of almost all recent research into code summarization is the attentional encoder-decoder neural architecture.  The inspiration for using models of this architecture derives from machine translation in NLP, in which sentences in one natural language (e.g. French) are translated into another (e.g. English).  When provided sufficient training data (usually well into the millions of examples), the encoder portion of the model learns a representation of one language, and the decoder learns the other.  The representations are combined via an attention network or other mechanism.  Then if the encoder is provided a sentence in one language, the decoder can be used to help predict an output sentence in the other language.  This is a tidy solution for machine translation because the information needed to write a sentence in one language tends to exist in translated sentences in other languages -- the encoder usually has access to all the information it needs to represent the sentence for the decoder.

Yet the metaphor of code summarization as machine translation breaks down because the source code of a subroutine is a very different representation of program behavior than natural language (an issue known as the concept assignment problem~\cite{biggerstaff1993concept}).  In a vast majority of neural code summarization approaches, the entire subroutine code is treated as a single unit, from which a summary is then written.  For example, the code may be a sequence entered into a Transformer or RNN-based encoder~\cite{hu2018deep, leclair2019neural, ahmad2020transformer}.  Or the Abstract Syntax Tree (AST) of the subroutine may be used as the input to a Graph Neural Network (GNN)~\cite{leclair2020improved} or a path-based approach~\cite{alon2019code2seq}.  The point is that the encoder is tasked with learning the semantics of an entire subroutine all at once.

A more human-like view of code is as a sequence of statements~\cite{robson1991approaches, siegmund2014understanding, siegmund2016program, latoza2007program}.  Software engineering research literature has long recognized that when people cannot just apply domain knowledge to rapidly digest code, ``they use bottom-up comprehension, so they understand source code statement by statement''~\cite{ siegmund2014understanding}.  The idea is that the behavior of any given statement tends to be highly dependent on the statements that precede it.  A conditional may cause a statement not to execute at all, or an assignment statement may completely change the meaning of a variable used later.  People understand code in terms of what each statement does.

The traditional way to establish the behavior of code in terms of the statements is dynamic analysis.  If actual inputs are entered and the code run, then the behavior of the program may be explained definitively.  Other strategies include symbolic or concolic execution~\cite{baldoni2018survey}.  But these strategies are not practical on a large dataset -- training data for neural code summarization often run into hundreds of thousands or even millions of examples~\cite{leclair2019recommendations, allamanis2019adverse}.

In this paper, we present \textbf{Statement-based Memory}, a novel encoder for neural source code summarization.  Our idea is inspired by Dynamic Memory Networks (DMN)~\cite{kumar2016ask} in which a model learns to answer questions by connecting features in a sequence of events (e.g., given two sentences ``Alice gave the ball to Bob.  Bob went to the parlor.'' the model could answer ``In what room is the ball?'' by learning to connect the feature ``Bob.'')  From a very high level, each statement is like a fact, and the features connecting the statements are the variable names, function calls, etc.  We create a novel model based on this inspiration using a custom attention mechanism among features in a sequence of statements.  The model learns to build a representation of each statement that emphasizes the important features for different concepts.  This learned attention serves as the model's ``memories.''

Our evaluation has three components: 1) we perform a quantitative controlled experiment using automated metrics, 2) we study the subset of subroutines most improved by our approach 3) we perform an analysis of our design choices and model configurations. The quantitative experiment maintains consistency with recent research literature by comparing our approach against a recent baseline over a large number of subroutines . The subset analysis serves to provide an in-depth view of the features of subroutines most improved by our approach. Finally, the evaluation of different configurations and design choices serves to maximize the breadth of our evaluation and inform future research.

We release all implementation details, experimental materials, and results via an online appendix in Section~\ref{sec:repo}.

\vspace{-0.1cm}
\section{Background \& Related Work}
\label{sec:background}
This section discusses key background technologies and related work, such as encoder-decoder models, source code summarization, and memory networks.

\vspace{-0.1cm}
\subsection{Source Code Summarization}
\label{sub:codesummary}

Automated source code summarization has interested the software engineering community for decades. Early approaches used Information Retrieval based techniques~\cite{sridhara2011automatically,mcburney2014automatic,rodeghero2014improving}. The neural machine translation (NMT) problem served as an early inspiration for a shift towards neural code summarization. Since 2014,  encoder-decoder models have formed the state-of-the-art in NMT as documented in the survey by Cho \textit{et al.}~\cite{cho2014properties}. A typical encoder-decoder model takes an input to the encoder in one language, for example French and translates the phrase into another language, such as English. For source code summarization, the input is source code of a subroutine and the output is a natural language summary. Figure~\ref{tab:screlated} provides an overview of the state-of-the-art in neural source code summarization over the last five years. These papers can be broadly classified into four categories based on the information encoded by the neural model:

\begin{figure}[!b]
	{\small
		\vspace{-0.4cm}
		\begin{tabular}{p{3.9cm}p{0.4cm}p{0.4cm}p{0.4cm}p{0.4cm}p{0.4cm}}
			 &T          &A          &C   &S                \\
			\textcolor{white}{*}Loyola~\emph{et al.}~(2017)~\cite{loyola2017neural}					& x	&   &   &   \\
			\textcolor{white}{*}Lu~\emph{et al.}~(2017)~\cite{lu2017learning}						& x	&   &   &   \\
			\textcolor{white}{*}Jiang~\emph{et al.}~(2017)~\cite{jiang2017automatically}			        & x	&   &   &   \\
			\textcolor{white}{*}Hu~\emph{et al.}~(2018)~\cite{hu2018summarizing}					& x	&x   &   &   \\
			\textcolor{white}{*}Hu~\emph{et al.}~(2018)~\cite{hu2018deep}							& x	&   &   &   \\
			\textcolor{white}{*}Allamanis~\emph{et al.}~(2018)~\cite{allamanis2018learning}			& x	&x   &   &   \\
			\textcolor{white}{*}Wan~\emph{et al.}~(2018)~\cite{wan2018improving}					& x	&   &   &   \\
			\textcolor{white}{*}Liang~\emph{et al.}~(2018)~\cite{liang2018automatic}					& x	&   &   &   \\
			\textcolor{white}{*}Alon~\emph{et al.}~(2019)~\cite{alon2019code2seq, alon2019code2vec}	& x	&x   &   &   \\
			\textcolor{white}{*}Gao~\emph{et al.}~(2019)~\cite{gao2019neural}						& x	&   &   &   \\
			\textcolor{white}{*}LeClair~\emph{et al.}~(2019)~\cite{leclair2019neural}					& x	&x   &   &   \\
			\textcolor{white}{*}Nie~\emph{et al.}~(2019)~\cite{nie2019framework}					& x	&   &   &   \\
			\textcolor{white}{*}Haldar~\emph{et al.}~(2020)~\cite{haldar2020multi}					& x	&x   &   &   \\
			\textcolor{white}{*}Ahmad~\emph{et al.}~(2020)~\cite{ahmad2020transformer}				& x 	&   &   &x   \\
			\textcolor{white}{*}Haque~\emph{et al.}~(2020)~\cite{haque2020improved}				& x	&   &x   &  \\ 
			\textcolor{white}{*}LeCLair~\emph{et al.}~(2020)~\cite{leclair2020improved}				& x	&x &    & \\
			\textcolor{white}{*}Feng~\emph{et al.}~(2020)~\cite{feng2020codebert}					& x	&   &    &x\\
			\textcolor{white}{*}Bansal~\emph{et al.}~(2021)~\cite{bansal2021project}					& x	&   &x  &  \\ 
			\textcolor{white}{*}Z{\"u}gner~\emph{et al.}~(2021)~\cite{zugner2021languageagnostic}		& x	&   &   &   \\
			\textcolor{white}{*}Liu~\emph{et al.}~(2021)~\cite{liu2021retrievalaugmented}				& x	&x   &   &   \\
			\textcolor{white}{*}(This Paper) 													& x	&   &   & x   \\
		\end{tabular}
	}
	\vspace{-0.1cm}
	\caption{Snapshot of the past five years in source code summarization. Column  $T$ means use of source code as Text. $A$ signifies learning from AST.  $C$ implies learning chiefly from code Context. $S$ means Self-Attention based approach}
	\label{tab:screlated}
\end{figure}

\textbf{Source Code as Text:} The earliest neural approach in literature is to input source code as text to an encoder-decoder model. These approaches typically use a Recurrent Neural Network (RNN), Transformer, or other neural networks to encode source code as if it were one long sequence. Then, the decoder translates that sequence into a natural language summary. In 2019 LeClair \textit{et al.}~\cite{leclair2019neural} introduced \textit{attendgru}, an implementation inspired from seq2seq model first proposed in 2014 by Bahdanau \textit{et al.}~\cite{bahdanau2014neural} for general purpose NMT. They observe that attention mechanism helps find important words in source code and generates better natural language summaries. However, these approaches are designed to learn common text sequence. Source code contains structure, data flow, and other information that this approach does not capture. We use this as a starting point for our approach in this paper.

\textbf{Abstract Syntax Tree (AST):} Abstract syntax trees are structural representations of source code. ASTs represent structure instead of text specific to a given piece of code, such as variables. Therefore, AST is a more generalizable representation of source code. In 2018 Alon \textit{et al.}~\cite{alon2018code2seq} introduced \textit{code2seq} that learns a structured representation of source code for summarization using AST paths. They observe that generalizability helps improve larger number of summaries overall . Then in 2020, LeClair \textit{el al.}~\cite{leclair2020improved} presented ~\textit{codegnngru} a bi-directional graph neural network (GNN) based approach. They observe that AST and source code can be attended to separately to provide orthogonal improvement. Codegnngru represents a family of GNN based techniques ~\cite{zhou2022automatic}. AST works well for simple and generalizable subroutines, but longer more complex subroutines may benefit from additional knowledge.

\textbf{External Context:} External context is information from sources other than the piece of code being summarized. Humans use external context during program comprehension~\cite{latoza2006maintaining}, so researchers are motivated to build models that use external context as well. In 2020, Haque~\textit{et~al.}~\cite{haque2020improved} introduced ``file context''  that encodes the file with the target source code. They use an attention mechanism to learn from other methods in the same file as the subroutine being summarized. Then in 2021, Bansal \textit{et al.}~\cite{bansal2021project} introduced ``project context'' extending upon the concept of file context. Their model encodes files and specific methods inside those files to add to the knowledge base. Although they show there is important information in context, the data and resource requirements are very high.

\textbf{Self-Attention:} Self-Attention techniques learn dependencies among tokens in the subroutine. Self-Attention techniques are hypothesized to learn structure without explicitly parsing AST, for example by Ahmad ~\textit{et al.}~\cite{ahmad2020transformer}. Essentially, self-attention mechanisms learn relationships between different parts of the input itself. Transformer~\cite{vaswani2017attention} models are a popular example of self-attention networks. In 2020 Ahmad \textit{et al.}~\cite{ahmad2020transformer} proposed a Transformer based model, that uses self-attention mechanism to summarize source code. On a larger scale Feng \emph{et al.}~\cite{feng2020codebert} introduce \textit{codeBERT}, an implementation of BERT~\cite{devlin2019bert}. They use multi-headed self attention to learn representation of source code for several code intelligence tasks. Resource and data requirements for CodeBERT are huge compared to other recent approaches. However, there are other lower cost self-attention techniques in literature. One of these is memory networks that we discuss in the next sub-section.

\vspace{-0.4cm}
\subsection{Memory Networks}
\label{sub:memorybackground}
\vspace{-0.15cm}
A memory network is a neural architecture that links information in different chunks of data that the network sees in sequence. These chunks of data are often referred to as \textit{facts}. The model sees the input as a sequence of facts and then learns connections among them. They are linked together by a series of learned attention units called \textit{memories} that are generated iteratively. Each memory represents the most important fact in that iteration. Each iteration attends to the memory from previous iteration and each of the facts. This allows the network to pick the most important fact from the series of memories, then the second most important fact given the previous fact, and so forth.

In 2015 Bordes \textit{et al.}~\cite{bordes2015large} introduce memory networks for a question answering system. The initial input to their system is a paragraph, where each sentence is a fact. Their approach stores an indexed array of facts they call the memory. In their approach , the memory vector gets updated every time a fact is added to the network. Similar facts are grouped together in memory to populate a hypergraph. Their approach attained improvements over the state-of-the-art approaches that append facts to the original sequence. 

Then in 2016, Kumar \textit{et al.}~\cite{kumar2016ask} introduced Dynamic Memory Networks (DMN). The input to their approach is a series of sentences, each one a fact. Their model is trained to learn weights for each sentence given a question-answer pair and is informed by the previous sentence. This self-attention mechanism is applied between the question and hidden states of all the sentences assigning weights to the most relevant ones iteratively. They observe that self-attention is beneficial to preserving temporal information in sequence of events. Their approach improves over the state-of-the-art at the time by linking temporal events together. This paper serves as an inspiration for our memory network. Application of memory networks is not limited to NLP. In literature, the benefits of applying memory networks can be seen across various domains .~\cite{yang2018learning,zhu2019cvpr}.

\section{Approach}
\label{sec:approach}

This section describes the approach of our paper. We list our model design, mathematical notation, parameters, and other details needed to replicate our proposed memory network.
\vspace{-0.2cm}
\subsection{Overview}
\label{sub:overview}
\vspace{-0.1cm}

\begin{figure}[!b]
	\vspace{-0.4cm}
	\centering
	\includegraphics[width=0.48\textwidth]{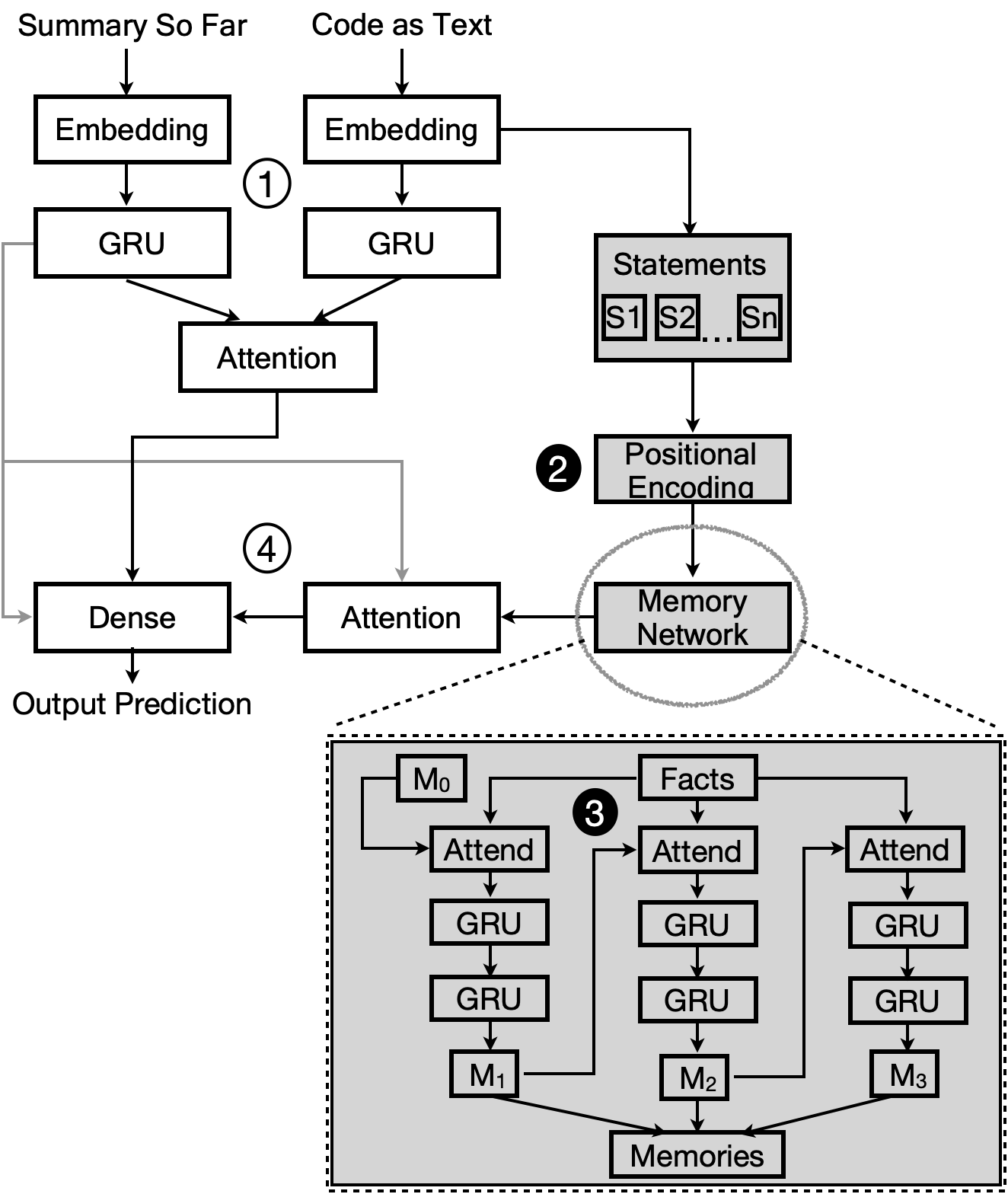}
	\vspace{-0.2cm}
	\caption{Overview of our model.}
	\label{fig:overview}
\end{figure}

Figure ~\ref{fig:overview} provides a high-level overview model. Our model accepts two inputs: 1) source code as text, and 2) summary sequence predicted so far. The model produces one output i.e. a matrix indicating the most likely candidate for the next word. Figure ~\ref{fig:overview} Area 1 follows the same architecture as ~\textit{attendgru} model described in Section~\ref{sub:codesummary}. We use the same embedding layer for the memory network because the tokens i.e. words in the sequences are exactly the same.

The novel contribution of this paper is the grey part in Figure ~\ref{fig:overview} area 2 and 3. First, we divide the source code of the java method into statements. Each statement is a line of code as it appears in the program. These serve as facts to our memory network. Second, we faithfully replicate positional encoding technique as defined by Sukhbaatar~\textit{et al.}~\cite{sukhbaatar2015end}. This encodes temporal information into the sequence i.e. the order in which the statements occur in the initial sequence. It is necessary to explicitly encode this information within each statement vector because the memory network attends to one statement at a time.

The memory network learns which of the statements are most important to predict a given word. This is achieved using a gated self-attention unit iteratively.  Figure ~\ref{fig:overview}  area 3 shows the inner workings of dynamic network. Over the first iteration, the learned attention unit i.e. the memory represents the most important statement. Over the subsequent iterations, it attends to the statements of secondary importance given high attention to the statement in previous memories. The number of iterations is a hardcoded value for all samples regardless of the number of statements.

As seen in Figure ~\ref{fig:overview} area 4, we use an attention mechanism twice. First, between the output of GRU encoder and the summary. Second, between the memories from the memory network and the summary. Attention mechanism helps the model learn parts of both GRU encoder and memories that are most important to the decoder. Finally we concatenate the outputs from these attention layers as well as the output of the summary GRU from Figure ~\ref{fig:overview} area 1. This large vector is passed to a dense layer used to predict the next word.

\vspace{-0.1cm}
\subsection{Model Details}
\label{sec:details}
\vspace{-0.1cm}
We explain the details of our model with the equations below. These details cover the key novel aspects to our model in Figure ~\ref{fig:overview} area 2. There are three steps. First, we split the sequence into statements, a process we describe in the next subsection. Second, we feed these statements to the positional encoder represented by these equations:
\vspace{-0.1cm}
\[\Big[\Big[P_{xy} = (1- y/Y) - (x/X)\times(1 - 2\times y/Y)\Big]_{x=0}^X\Big]_{y=0}^Y\]
\[F= S \otimes P' \]

Here $P$ is the positional encoding matrix, $S$ is the vector of statements, $X$ is the size of the word embedding, and $Y$ is the maximum length of a statement. The $ [  ]_{x=0}^X$ notation describes a \textit{for loop} from $0$ to $X$. The $\otimes$ symbol represents element wise multiplication. $P'$ indicates transpose of the positional encoding matrix. Third, we present the memory network layer that receives a set of statements from the model. We represent our memory network with these equations:
\vspace{-0.1cm}
\[{\Big[Memories_i= Update(F,Q,M_{(i-1)})}\Big]_{i=0}^h  \tag{1}\label{eq:1}\]

Here $h$ is the number of ``hobs'', i.e. pre-defined iterations the memory network goes through and $Q$ is a matrix where 0.1 is numerical the value of each element . The function ``Update'' can be represented by the following equation:
\vspace{-0.1cm}
\[{\Big[m = G'_t \times GRU(F_t ,m) + (1-G'_t)\times m}\Big]_{t=0}^n\]

Here $m$ is initialized a matrix of zeros. After iterating through the attention for all statements using a GRU, the final $m$ matrix is returned as an output to the Update function i.e. $Memories_i$ from equation~\ref{eq:1}. $G'_t$ is output of the gated attention unit for that fact, represented by equations:
\vspace{-0.1cm}
\[G_k = (F_k \times Q_k) \oplus (F_k \times M_k) \oplus |F_k-Q_k| \oplus |F_k - A_k|\]
\[G'_k = \Sigma (tanh(G_k))\]

Here, $\oplus$ represents the tensor concatenation operation and $G'_k$ is the gated attention learned between statement k($F_k$) and memory k($M_k$). This is iterated via a for loop from $0$ to $n$, where $n$ is the maximum number of statements. The $\Sigma$ and $tanh$ computations are described in the Keras library, version details for which are in Section~\ref{sec:versions}.

At the decoder side seen in Figure ~\ref{fig:overview} area 4, the output memories vector is concatenated with the output of the summary GRU in Figure ~\ref{fig:overview} area 1. This concatenated vector is used to compute the attention using softmax activations. The attention vector is then applied back to the memories vector using a dot product. This approach is similar and in addition to the attention mechanism in the \textit{attendgru} encoder. In the final step we concatenate both post-attention vectors as well as the output from the summary GRU into a \textit{context} vector. A keras dense layer with the softmax function takes this vector and produces the output word.

\vspace{-0.1cm}
\subsection{Input\slash Output}
\label{sec:io}
\vspace{-0.1cm}

Our model accepts two inputs: 1) the method source code as tokenized text and 2) the summary predicted so far as tokenized text. This tokenization occurs during pre-processing. Input 1 also contains ``<NL> '' tokens which mark the end of a line of code. We partition code at these tokens to extract statements ensuring each statement is a vector representing a line of code and the statements are kept in order to form a $m \times n$ matrix, where $m$ is the number of statements and $n$ is the maximum number of words in a statement. For our experiment we use $m=70$ and $n=30$. This design configuration is explored further in Section~{sec:quantexp}. 

Our model outputs a $v \times v$ matrix where $v$ is the size of tokenized vocabulary for summaries. The max value in this vector is selected as the word added to the previously generated sequence, for the next iteration during testing. For the Java dataset, vocabulary size $t$ for source code is 69725 and vocabulary size $v$ for summaries is 10908. For the Python dataset  vocabulary size $t$ for source code is 100000 and vocabulary size $v$ for summaries is 25000.  These limits on vocabulary strike a delicate balance between limiting resource exhaustion and representing the most common words. The model sequentially generates a maximum of 12 tokens for a total summary length of 13 words where the \textit{<\slash s>} token marks the end of the generated summary.

\vspace{-0.15cm}
\subsection{Combining with non-statement encoders }
\label{sec:ensembles}
\vspace{-0.05cm}

We combine our statement based memory with non statement encoders from recent literature. We use an ensemble technique as recommended for code summarization by LeClair~\textit{et al.}~\cite{leclair2021ensemble}. Ensemble of models is used in a wide range of research areas such as Gene Identification ~\cite{le2020computational} and Computer vision~\cite{qummar2019deep} . The core reason ensembles are popular is that they take advantage of two or more approaches where each one might excel at a niche use case. LeClair~\textit{et al.} found that ensembles are able to capture and merge orthogonal contributions of various approaches in source code summarization. We follow best practices laid out in their paper to create ensemble models for source code summarization.

We created ensemble models by combining a non-statement based model and a statement based approach. We follow the methods listed in  LeClair ~\textit{et al.} and choose a "mean" aggregation method, where softmax output from both approaches is averaged to generate the output word. We choose this method of aggregating ensembles because it is simpler than other methods and easier to standardize across different encoders. More importantly, creating ensembles this way has negligible resource overhead because these trained models can be reused to create any number of ensembles. Every model we used to create ensembles was trained independently and only aggregated during prediction. For example, {\small \texttt{transformer+SMN}} uses {\small \texttt{SMNcode }} i.e. a model trained with our statement based memory network encoder as described in Section~\ref{sub:overview} ,  and {\small \texttt{transformer}} a model trained with a transformer based encoder (see Section ~\ref{sec:baselines}). 
 
 \vspace{-0.2cm}
\subsection{Hyperparameters}
\label{sec:hyper}
Our system is designed with the following hyperparameters:

\begin{table}[h!]
	\vspace{-0.3cm}
	\centering
	\begin{tabular}{lll}
		$tdatlen$	&200	  & maximum length of code sequence\\
		$comlen$	&13	& maximum length of summary\\
		$e_{dim}$ & 100	& dimension of word embedding\\
		$l_{dim}$  &100	& dimension of encoding layers\\
		$h$		& 3	& iterations for memory self-attention \\
		$n$   	& 70    & maximum number of statements \\
		$Y$  	& 30   & maximum length of a statement \\
		$RNN$ 	& GRU   & type of RNN \\
		$b$	& 100	& batch size \\
		$p$		&Adam		& parameter optimization technique \\
		$l$		&cross-entropy		& loss function\\
	\end{tabular}
	\vspace{-0.3cm}
\end{table}

\vspace{-0.3cm}
\subsection{Hardware\slash Software Versions}
\label{sec:versions}
Our models have been trained and tested on the following Hardware and Software versions :

\textbf{Hardware:} For training, validation and testing we used a single TITAN RTX GPU with 24 GBs of memory , and Intel(R) Xeon(R) CPU E5-1620 v3 @ 3.50GHz CPU.

\textbf{Software:} We used these software versions: Ubuntu LTS 18.04 , CUDA 11,  Tensorflow 2.4.1,  Python 3.6.9, Keras 2.4.0, NLTK 3.6.2, numpy 1.19.5, scipy 1.7.3,  and sklearn 1.0.2

\section{Dataset Preparation}
\label{sec:dataset}
We used two datasets specifically curated for our paper : 1) Java dataset, and 2) Python dataset.

The Java dataset we use consists of 190K Java methods that we curated from a larger dataset of 2.1m methods proposed in 2019 by LeClair \textit{el al.}~\cite{leclair2019recommendations} named \textit{funcom}. We selected the ~\textit{funcom} dataset for three reasons. First, the summaries were generated with the help of JavaDocs, with high quality standardized documentation. Second, all subroutines in the dataset are split at project-level not at method level to reduce the risk of  data leaks. Third, the dataset is well established and follows best practices from literature~\cite{leclair2019neural,allamanis2018learning}. 

We curated our dataset by first removing code clones from the dataset as proposed by Bansal~\textit{et al.}~\cite{bansal2021project} using the technique presented by Microsoft Research ~\cite{allamanis2019adverse}. We also select 190k of the longest methods from the filtered dataset. This reduces the likelihood of smaller get/set/return methods with generic summaries. The \textit{funcom} dataset contains a large number of these methods as observed by Haque et al.~\cite{haque2021action}. They observed that it was comparatively easier to generate summaries that would achieve high metric scores for these methods. This is due to the distinct structure and limited number of action words describing the purpose of the method. Longer methods also make better candidates for statement-based memory networks as the network is designed to learn attention between multiple statements.

We created the Python dataset using programs we extracted from Github. To keep the dataset comparable with Java, we apply the same filters to remove duplicates and smaller methods with limited number of sentences. The Python dataset consists of 270k total python functions. We selected the 270k largest functions from a dataset of 1.7 million filtered python functions. These functions are split at the project-level to reduce the risk of data leaks following the same best practices used for the Java dataset. 

\vspace{-0.15cm}
\section{Experiment Design}
\label{experiment}
In this section we describe our experimental design including the research questions, methodology, baselines, and metrics used to evaluate our model against the baseline.
\vspace{-0.25cm}
\subsection{Research Questions}
\label{sec:rqs}
 Our main research objective is to evaluate if statement based memory network (SMN) improves source code summarization. Also, we aim to learn about the subset of summaries most improved by our approach and how our design choices affect performance. To that end, we ask three research questions:

\begin{description}
	\item[RQ1] What is the difference between our approach and the baseline when ensembled with non-statement-based models as measured by automated metrics?
	\item[RQ2] What are the features of the subset of summaries most improved by our approach?
	\item[RQ3] How do the different design choices and configurations affect the performance of our encoder, as measured by automated metrics?
	\end{description}

The rationale behind RQ1 is to evaluate our approach in a way consistent with previous approaches. We designed our encoder to add to, not compete with existing approaches. To that end, a vast majority of papers in the past 5 years use ensembles to evaluate orthogonal improvements by combining different approaches ~\cite{leclair2021ensemble,zhou2022summarizing,gao2021code}.  We use automated metrics to quantify the differences between our approach and the baseline, consistent with literature ~\cite{haque2022semantic,roy2021reassessing}.
 
The rationale for RQ2 is to further investigate the impact of our encoder in improving a subset of summaries when compared against the basic models. While we designed our approach to add to, and not compete with other approaches, competitive comparison helps place our approach among the state-of-the-art. Non-ensembled evaluation helps highlight the subset of summaries most improved by our approach.

The rationale behind RQ3 is that a neural network is a complex system and several factors may impact performance. We make several design choices informed by literature to construct our model. It could be useful to know which of these factors are most crucial to the performance of our encoder.   We evaluate three different design choices against alternatives informed by best practices in literature to aid any future work that may benefit from our approach. We test these configurations for our encoder without ensembles because minor differences can be overshadowed by ensembling.

\vspace{-0.2cm}
\subsection{Methodology}
\label{sec:methodology}
\vspace{-0.1cm}

Our methodology to answer RQ1 is informed by both the procedure established by most source code summarization approaches ~\cite{hu2018deep,alon2019code2seq,leclair2019neural} and the latest research that suggests improvements~\cite{haque2022semantic,roy2021reassessing,9286030}. We start with an established dataset of subroutine-summary pairs and curate it for our experiment (see~Section~\ref{sec:dataset}). We then train models of each of the non-statement based basic models and our statement based approach as well as another statement-based baseline. These models are independently trained till convergence. Convergence is popularly accepted as the epoch at which validation accuracy peaks without substantially higher validation loss. 

Minor performance improvements can be achieved by simply making the model bigger, i.e. increasing the dimensions of the decoder vectors. Consequently, we keep the size of all embedding and dense layers constant. The penultimate step is to ensemble each combination of statement based and non-statement based models using techniques from literature~\cite{leclair2021ensemble}. Lastly, we use automated metrics for evaluation. Although the BLEU metric has been popular in the past, recent studies suggest BLEU scores may not accurately reflect an improvement in quality of summaries~\cite{wieting-etal-2019-beyond,haque2022semantic,roy2021reassessing}. Therefore, we also evaluate using METEOR, recommended by Roy~\textit{et al.}~\cite{roy2021reassessing} and USE, recommended by Haque~\textit{et al.}~\cite{haque2022semantic}.  

To aid the evaluation of our ensembles, we also discuss the features of the ``difference set'' between ensemble with our approach and the baseline statement-based approach.  The difference set is the set of predicted summaries which are different for any two approaches.  Typically, there exists a subset of generic or otherwise very easy to summarize subroutines, which a majority of neural models will summarize in exactly the same way . This group of easy subroutines is the ``same set''  that may exaggerate the overall scores to a degree that may distort the reported results.

We use automated metrics for our RQs in favor of a human study for three reasons. First, we have several thousands of predictions to test against several baselines, and configurations. A small subset that a human participant could reasonably be expected to evaluate would not be representative of the performance of our approach. Second, human studies tend not to be reproducible because they are opinion-based. Any results obtained may be specific to the demographic of participants recruited. Third, it is expensive to conduct a reliable human study that would require a large number of participants. Cost-decreasing methods lead to unreliable evaluations by either recruiting participants without enough experience with source code or a very small number of participants with low agreement scores within their ratings.

To answer RQ2 we evaluate our approach in two ways. First, we evaluate our model against each of the other baselines without the use of ensembles and present the results using the same automated metrics as RQ1. Second, we report the features of the ``improved set'', a subset where our approach achieves significantly higher score compared to the basic models. This subset is obtained by taking each pair of predicted summary and comparing to the reference on the basis of scores alone. Note that this is not the same as the difference set in RQ1, which considers differences between the words in the summary. We only use METEOR scores because: 1) BLEU is only reliable at the corpus level, and 2) USE is volatile because of the learnt weights that are used to create the vectors.

For RQ3  we conduct three tests for different configurations of our encoder. Each test is of one design choice: 1) $h$ iterations for dynamic network, 2) randomly initialized matrix $Q$, and 3) Positional encoding. We compare alternatives to these design choices against the best performing model configuration described in Section~\ref{sec:approach}. For the first test, we change  $h$ that is the number of iterations used to generate memories, described  in Section~\ref{sec:details}. We compare models H1, H2, H4, and H5 trained with four different $h$ values against our optimal configuration. For the second test, we replace the $Q$ matrix (recall Equation \ref{eq:1} in Section~\ref{sec:details}) with the vector representing summary generated so far, a configuration inspired by Kumar~\textit{et al.}~\cite{kumar2016ask}  we call this configuration {\small \texttt{SMN-summary vector}} . This is in contrast to the best performing model inspired by the works of Sukhbaatar~\textit{et al.}~\cite{sukhbaatar2015end}. For the third test, we compare End of Sentence (EOS) encoding as described by Kumar~\textit{et al.}~\cite{kumar2016ask} against positional encoding used in our best configuration described in Section~\ref{sec:details}.  

\vspace{-0.2cm}
\subsection{Metrics}
\label{sec:metrics}
We use three metrics for evaluation of our model. First, BLEU~\cite{papineni2002bleu}, a natural language similarity metric that computes n-gram overlap. Second, METEOR~\cite{banerjee2005meteor} uses unigram precision and recall and computes the harmonic mean. Third and latest, USE~\cite{cer2018universal}, which encodes summaries into a vectors and computes the distance between predicted and reference summary vectors. Each of these metrics is backed by human studies to ensure the metrics correspond to human perception. 

Recent study by Haque~\textit{et al.}~\cite{haque2022semantic} compares these metrics for source code summarization. Their study suggests unigram or n-gram approaches may not generalize to source code summarization jargon as well as they do to natural language summaries. We follow their recommendation to evaluate generated summaries with USE. We use the augmented implementation of USE provided as part of their replication package. We follow recommendations by Roy ~\textit{et al.}~\cite{roy2021reassessing} to use METEOR and Graham~\textit{et al.} ~\cite{graham2014randomized} to compute statistical significance of the USE and METEOR score difference.

\vspace{-0.2cm}
\subsection{Basic Models : Non-Statement-based}
\label{sec:baselines}
Our basic models consist of four recent neural network based approaches in source code summarization literature that do not use statement based information. We faithfully re-implement each of these approaches, taking the novel contributions of their approach (mostly encoder) and incorporate them within our sequence to sequence framework:

\begin{description}
	\item[code2seq] is the approach proposed by Alon ~\textit{et al.}~\cite{alon2019code2seq} that represents a family of papers that use the AST to represent source code. They use AST path tokens as input and encode this information using multiple GRU. They use a decoder similar to the \textit{attendgru} baseline. 
	\item[attendgru] was proposed by LeClair ~\textit{et al.} ~\cite{leclair2019neural} as a typical sequence to sequence model. They use source code as a sequence of text tokens as input . They process these inputs with a learnt word embedding and use GRUs to encode the inputs and a GRU for the decoder.
	\item[transformer] proposed by Ahmad \textit{et al.}~\cite{ahmad2020transformer} represents a family of transformer based approaches . They use source code text tokens as input and the use multi-headed self attention between all tokens in the input. We faithfully re-implement the model within our framework.
	\item[codegnngru] by LeClair ~\textit{et al.}~\cite{leclair2020improved} is a representative example of graph neural networks for source code summarization. They encode both AST and source code as text in form of a graph. They use the nodes and adjacency list for this graph as input. They present several configurations in their paper. We re-implement the GRU based approach for posterity from the replication package provided. 

\end{description}

This is not a complete list of recent basic non-statement based models, but each of them is representative of a family of approaches as explained in Section~\ref{sub:codesummary}. We do not replicate a basic model to represent external context because those techniques benefit from additional data that our framework does not have such as UML~\cite{wang2021cocosum} or an indexed corpus~\cite{li2021editsum}. We considered using CodeBERT by Feng \emph{et al.}~\cite{feng2020codebert} as a basic approach. We did not replicate their approach because: 1) the resource requirements for their approach are huge and would take us months to train a replication of the pre-trained embedding presented, and 2) the improvements for source code summarization is similar to those reported by other baselines. Therefore we decided not to replicate their approach in favor of other low resource approaches.

\subsection{Baseline: Hierarchical Attention Network}
\label{han}
Our main baseline is {\small \texttt{HANcode}}, a statement based approach by Zhou {et al.}~\cite{zhou2022summarizing}. It was released in early 2022 and is the most recent approach that uses similar techniques to our approach such as intra-attention and statements-flow. We chose HANcode as our main competitor because the paper establishes the benefits of using statement-based approach. We design our experiment to evaluate our hypothesis that memory networks are better at encoding statement-level information.

HANcode is inspired by the Hierarchical Attention Network (HAN) proposed for document classification by Yang~\textit{et al.}~\cite{yang2016hierarchical}. Their approach is a statement based approach like ours, however applies word level self-attention much like the {\small \texttt{transformer}} baseline. We faithfully replicate their approach within our own framework using GRUs. We ensemble this approach with the four non-statement-based approaches to serve as baseline against which we evaluate our statement-based memory network. 
\vspace{-0.1cm}
\subsection{Threats to Validity}
\label{sec:threats}
The major threats to validity of our study are: 1) limitations of our datasets 2) metrics used to evaluate performance, and 3) the hyperparameters and configurations. 

First, although we curate the Java dataset from a well-established dataset in the field, similar results may not be seen over other datasets. For example, datasets that mainly consists of very small subroutines with one or two statements, may not see any improvement from a statement based approach like ours. Datasets in other programming languages may not have information embedded between statements or may need different model configurations. To mitigate this our dataset has been through multiple rounds of filters to create a high quality subset. We created the Python dataset that has similar features as the Java dataset to further mitigate this.

Second, automated metrics like METEOR and BLEU have been established in NMT. However, they only provide a comparison to the reference summary, where a high score means the generated summary has mostly the same words as the reference. This does not take into consideration that some words in the summary may be more important than others. This is specially true to technical jargon one would expect to see in a code summary. To mitigate these challenges we evaluate using the latest USE metric and follow the most recent guidelines for using NMT metrics for source summarization~\cite{roy2021reassessing,haque2022semantic}.

Third, implementation configurations are another threat to validity. It is possible to get minor improvements in performance by tweaking hyperparameters and increasing the size of the model. Therefore we replicate each baseline very carefully ensuring similar dimensions for all vectors involved. The ensemble process we used to combine our statement based encoder with a non-statement based model can distort the results. We attempt to mitigate this threat by individually comparing all the models we trained to create ensembles in RQ2 and then compare different configurations of our encoder without any basic models in RQ3.

\begin{table*}[t]
	\centering
\caption{Metric scores for SMN encoder ensembled with each basic models compared against HAN encoder ensembled with the same baselines. T-test is conducted between +SMN and +HAN to reject a null hypothesis with a P-value <0.05 }
\vspace{-0.2cm}
\begin{tabular}{lllllllllllll} 
 	\cline{2-12}
	& \multicolumn{1}{|p{2.5cm}|}{Java}                                 	& \multicolumn{4}{c|}{METEOR}                                                                                  			& \multicolumn{4}{c|}{USE}                                        			&\multicolumn{2}{c|}{BLEU}   	&  \\
	& \multicolumn{1}{|p{2.5cm}|}{\textbf{Basic}} 	&{+HAN}  		&{+SMN} 			& {T-test}        		&\multicolumn{1}{c|}{P-value}  	   	& {+HAN}		& {+SMN} 		& {T-test}   	&\multicolumn{1}{c|}{P-value}       	& {+HAN}		&\multicolumn{1}{c|}{+SMN} 	&  \\ \cline{2-12}
	& \multicolumn{1}{|l|}{code2seq}     			& {32.01}		& {35.82} 			& {23.60}   		&\multicolumn{1}{l|}{<0.01} 	    	& {47.01}		& {52.61}         		& {33.09}        	&\multicolumn{1}{l|}{<0.01}       		& {16.87}     	&\multicolumn{1}{l|}{19.95}  	&    \\
	& \multicolumn{1}{|l|}{attendgru}     			& \textbf{34.30}	& {36.16} 			& {12.20}   		&\multicolumn{1}{l|}{<0.01} 	    	& \textbf{50.59}	& {53.28}         		& {18.16}        	&\multicolumn{1}{l|}{<0.01}    	    	& \textbf{19.23}	&\multicolumn{1}{l|}{\textbf{20.52}}  	&    \\
	& \multicolumn{1}{|l|}{transformer}     		& {34.00}		& \textbf{36.19} 	& {14.42}   		&\multicolumn{1}{l|}{<0.01} 		& {50.45}		& \textbf{53.40}         	& {20.12}        	&\multicolumn{1}{l|}{<0.01}            	& {18.60}     	&\multicolumn{1}{l|}{20.31}  	&    \\
	& \multicolumn{1}{|l|}{codegnngru}     		& {33.40}		& {36.08} 			& {17.13}   		&\multicolumn{1}{l|}{<0.01} 	 	& {48.80}		& {52.91}          		& {26.10}        	&\multicolumn{1}{l|}{<0.01}	    	& {18.29}		&\multicolumn{1}{l|}{20.17}  	&    \\ \cline{2-12}
\end{tabular}

\hspace{0.01cm}
\begin{tabular}{lllllllllllll} 
 	\cline{2-12}
	& \multicolumn{1}{|p{2.5cm}|}{Python}                                 	& \multicolumn{4}{c|}{METEOR}                                                                                  			& \multicolumn{4}{c|}{USE}                                        			&\multicolumn{2}{c|}{BLEU}   	&  \\
	& \multicolumn{1}{|p{2.5cm}|}{\textbf{Basic}} 	&{+HAN}  		&{+SMN} 			& {T-test}        			&\multicolumn{1}{c|}{P-value}  	   	& {+HAN}		& {+SMN} 		& {T-test}   	&\multicolumn{1}{c|}{P-value}       	& {+HAN}		&\multicolumn{1}{c|}{+SMN} 	&  \\ \cline{2-12}
	& \multicolumn{1}{|l|}{code2seq}     			& {23.93}		& {27.14} 			& {14.89}   			&\multicolumn{1}{l|}{<0.01} 	    	& {37.39}		& {41.73}         		& {21.01}        	&\multicolumn{1}{l|}{<0.01}       		& {16.50}     	&\multicolumn{1}{l|}{19.61}  	&    \\
	& \multicolumn{1}{|l|}{attendgru}     			& {28.74}		& {29.76} 			& {05.94}   			&\multicolumn{1}{l|}{<0.01} 	    	& {42.57}		& {44.51}         		& {11.68}        	&\multicolumn{1}{l|}{<0.01}    	    	& {21.27}		&\multicolumn{1}{l|}{22.41}  	&    \\
	& \multicolumn{1}{|l|}{transformer}     		& \textbf{30.18}	& \textbf{30.8} 		& {{\color{white}0}3.82}   	&\multicolumn{1}{l|}{<0.01} 		&\textbf{43.49}	& \textbf{45.30}         	& {11.40}        	&\multicolumn{1}{l|}{<0.01}            	& \textbf{22.77}	&\multicolumn{1}{l|}{\textbf{23.30}}  	&    \\
	& \multicolumn{1}{|l|}{codegnngru}     		& {28.9}		& {29.79} 			& {{\color{white}0}5.19}   	&\multicolumn{1}{l|}{<0.01} 	 	& {41.76}		& {43.72}          		& {11.42}        	&\multicolumn{1}{l|}{<0.01}	    	& {22.08}		&\multicolumn{1}{l|}{23.12}  	&    \\ \cline{2-12}
\end{tabular}
\label{tab:rq1}
\vspace{-0.6cm}
\end{table*}
\begin{table*}[t]
	\vspace{0.3cm}
	\centering
	\caption{Features of the Difference set. Diff(\%) column contains the size of the difference set. Same column metric scores for the same set. +HAN and +SMN indicate metric scores over the difference set for ensemble of the basic models with the baseline and our approach respectively. T-test is conducted between the +SMN and +HAN scores over the difference set.}
	\vspace{-0.2cm}
\begin{tabular}{@{}lllllllllllllll|ll} 
 	\cline{2-16}
	& \multicolumn{1}{|p{1.25cm}|}{Java}                     &\multicolumn{1}{c|}{Diff}        						& \multicolumn{5}{c|}{METEOR}                                                                 					& \multicolumn{5}{c|}{USE}           													&\multicolumn{3}{c|}{BLEU}  	&  \\
	& \multicolumn{1}{|p{1.25cm}|}{\textbf{Basic}} &\multicolumn{1}{c|}{ (\%)}        		&{Same}		&{+HAN} 			&{+SMN } 	& {T-test}   	&\multicolumn{1}{c|}{P-val} 		& {Same} 		& {+HAN} 			& {+SMN }	& {T-test} 		&\multicolumn{1}{c|}{P-val}  	 & {Same}		&\multicolumn{1}{c}{+HAN}  		&\multicolumn{1}{c|}{+SMN } 	&  \\ \cline{2-16}
	& \multicolumn{1}{|l|}{code2seq}     			&\multicolumn{1}{l|}{87.13}		&{60.11}		&{27.86} 			&{32.24}		& {23.71}   	&\multicolumn{1}{l|}{<0.01}		&{71.93}		& {43.33}                	& {49.76} 		& {33.40} 		&\multicolumn{1}{l|}{<0.01}         &{49.17}		&\multicolumn{1}{c}{12.36}  		&\multicolumn{1}{l|}{15.49}    \\
	& \multicolumn{1}{|l|}{attendgru}     			&\multicolumn{1}{l|}{79.43}		&{56.58}		&{28.53} 			&{30.88}		& {12.23}   	&\multicolumn{1}{l|}{<0.01} 		&{69.36}		& {45.73}                 	& {49.12} 		& {18.25} 		&\multicolumn{1}{l|}{<0.01}  	&{46.60}		&\multicolumn{1}{c}{11.95}  		&\multicolumn{1}{l|}{12.39}    \\
	& \multicolumn{1}{|l|}{transformer}     		&\multicolumn{1}{l|}{80.91}		&{57.90}		&{28.36} 			&{31.06}		& {14.46}   	&\multicolumn{1}{l|}{<0.01} 		&{69.40}		& {45.98}                 	& {49.63} 		& {20.23}  		&\multicolumn{1}{l|}{<0.01}          &{48.23}		&\multicolumn{1}{c}{12.45}  		&\multicolumn{1}{l|}{13.65}    \\
	& \multicolumn{1}{|l|}{codegnngru}     		&\multicolumn{1}{l|}{82.67}		&{59.02}		&{28.03} 			&{31.28}		& {17.20}   	&\multicolumn{1}{l|}{<0.01} 		&{71.05}		& {44.14}                 	& {49.11} 		& {26.32}    	&\multicolumn{1}{l|}{<0.01}	&{48.62}		&\multicolumn{1}{c}{12.24}  		&\multicolumn{1}{l|}{12.24}    \\ \cline{2-16}
\end{tabular}
\begin{tabular}{@{}lllllllllllllll|ll} 
 	\cline{2-16}
	& \multicolumn{1}{|p{1.25cm}|}{Python}                         &\multicolumn{1}{c|}{Diff}        						& \multicolumn{5}{c|}{METEOR}                                                                 & \multicolumn{5}{c|}{USE}           								&\multicolumn{3}{c|}{BLEU}  	&  \\
	& \multicolumn{1}{|p{1.25cm}|}{\textbf{Basic}} &\multicolumn{1}{c|}{(\%)}        	&{Same}		&{+HAN} 			&{+SMN } 	& {T-test}   			&\multicolumn{1}{c|}{P-val} 		&{Same}			& {+HAN} 				& {+SMN }	& {T-test} 			&\multicolumn{1}{c|}{P-val}  	 	&{Same}    		&\multicolumn{1}{c}{+HAN}  		&\multicolumn{1}{c|}{+SMN } 	&  \\ \cline{2-16}
	& \multicolumn{1}{|l|}{code2seq}     			&\multicolumn{1}{l|}{90.15}	&{94.45}		&{16.23} 			&{19.79}		& {14.90}   			&\multicolumn{1}{l|}{<0.01}	&{95.73}		& {31.02}                	& {35.83} 		& {21.05} 		&\multicolumn{1}{l|}{<0.01}        &{93.49}       	&\multicolumn{1}{c}{7.97}  		&\multicolumn{1}{l|}{11.54}    \\
	& \multicolumn{1}{|l|}{attendgru}     			&\multicolumn{1}{l|}{84.98}	&{91.08}		&{17.72} 			&{18.92}		& {{\color{white}0}5.94}   	&\multicolumn{1}{l|}{<0.01} 	&{93.07}		& {33.64}                 	& {35.92} 		& {11.70} 		&\multicolumn{1}{l|}{<0.01}  	&{89.66}		&\multicolumn{1}{c}{8.71}  		&\multicolumn{1}{l|}{10.12}    \\
	& \multicolumn{1}{|l|}{transformer}     		&\multicolumn{1}{l|}{82.94}	&{89.28}		&{18.02} 			&{18.78}		& {{\color{white}0}3.82}   	&\multicolumn{1}{l|}{<0.01} 	&{92.00}		& {33.51}                 	& {35.69} 		& {11.41}  		&\multicolumn{1}{l|}{<0.01}        &{87.50}        	&\multicolumn{1}{c}{9.04}  		&\multicolumn{1}{l|}{{\color{white}0}9.56}    \\
	& \multicolumn{1}{|l|}{codegnngru}     		&\multicolumn{1}{l|}{82.97}	&{91.34}		&{16.09} 			&{17.16}		& {{\color{white}0}5.20}   	&\multicolumn{1}{l|}{<0.01} 	&{93.34}		& {31.18}                 	& {33.53} 		& {11.43}    	&\multicolumn{1}{l|}{<0.01}	&{90.06}		&\multicolumn{1}{c}{7.46}  		&\multicolumn{1}{l|}{{\color{white}0}8.73}    \\ \cline{2-16}
\end{tabular}
\vspace{0.1cm}
\label{tab:rq2}
\vspace{-0.3cm}
\end{table*}
\section{Experimental Results}
This section includes experimental results for all three of the RQs we ask in Section~\ref{sec:rqs}.

\subsection{RQ1: Comparison against HANcode baseline}
\label{sec:rq1}
We report our evaluation results in Table~\ref{tab:rq1}. The statistical tests are relative to models ensembled with HAN instead of SMN. We make three interesting observations for the Java dataset. First, every ensemble with {\small \texttt{SMNcode}}  achieves statistically significant performance gains when compared to the individual baseline models. Second, the highest performing ensemble for USE and METEOR scores is {\small \texttt{transformer+SMNcode}} slightly better than {\small \texttt{attendgru+SMNcode}}. We posit that {\small \texttt{transformer}} as a word-level self attention model complements our statement-level approach more so than {\small \texttt{attendgru}}~. 

For the Python dataset, in addition to the observations from ensembles over Java dataset, one observation stands out. The ensemble {\small \texttt{transformer+HAN}} achieves scores higher than {\small \texttt{attendgru+HAN}} model which performed best among HAN combination for the Java dataset. This may indicate that python dataset might specially benefit from token-level self-attention. We further evaluate the differences between the two datasets by comparing the difference set, as explained in Section~\ref{sec:methodology}.

In Table~\ref{tab:rq2} we present the features of the difference set between our basic models ensembled with {\small \texttt{SMNcode}} and the same models ensembled with {\small \texttt{HANcode}} for both datasets. Overall, we observe that our approach improves performance over a large number of summaries. For both datasets, we observe a large difference set. 

For the Java dataset, we observe that 79-87\% of the summaries are part of the difference set. We observe each difference set is improved by a statistically significant score for all three metrics when compared to the baseline. We found that the small set with same summaries for both approaches has a really high score for every model. One possible reason is that these summaries have descriptive names. For example, method name : DrawCartesianMap() has the reference summary ``Draws Cartesian Map''. These are relatively easy for any approach to get right. This makes the difference set a more reliable way to look under the hood and see where our approach really makes a difference. With a 10\% or higher score difference over the baseline HAN, we posit our memory network approach improves the state-of-the-art in statement based encoders.

For the Python dataset we observe that  82-90\% of the summaries are part of the difference set. We observe a similar trend to the Java dataset in terms of improvement in metric scores when compared to the baseline. We observe a three key differences : 1) a slightly larger difference set, 2) much higher score for the same set, and 3) lower delta and scores for the difference set. We created this dataset explicitly for this paper, so these observations are very important. First, the slightly larger difference set indicates that overall there are less number of functions that are easy to summarize. This makes sense as the Java summaries were generated by JavaDocs, but the Python summaries were extracted from various repositories, and thus are less likely to follow consistent naming conventions as observed for Java. Second, upon further investigation of the same set we found that the high scores on the same set are from functions with generic summaries like ``Takes input X and Y''. We split the dataset by project to reduce the likelihood of data leaks between training and testing sets. However, within the project some generic summaries tend to repeat for non-duplicate functions. Third, the delta between our approach and the baseline as well as overall scores are lower for the difference set. Although the overall scores in Table~\ref{tab:rq1} seem similar for Python and Java, difference set analysis reveals that Python dataset is more challenging. However, the SMN encoder improves over the baseline in each instance.

\begin{table*}[t]
	\centering
	\caption{Metric scores for non-ensembled basic models as well as our baseline HANcode. SMNcode is the standalone model for our statement based memory encoder. T-tests are conducted between each other model and the SMNcode model. }
	\vspace{-0.2cm}
\begin{tabular}{llllllllllll} 
 	\cline{2-9}
	& \multicolumn{1}{|p{3cm}|}{Java} 								& \multicolumn{3}{p{3cm}|}{METEOR}   							& \multicolumn{3}{p{3cm}|}{USE}                                        			&\multicolumn{1}{p{1cm}|}{BLEU}  	&  \\
	& \multicolumn{1}{|p{3cm}|}{\textbf{Models}}         	&{Score} 		& {T-test}   			&\multicolumn{1}{c|}{P-value}       		& {Score} 		& {T-test}        			&\multicolumn{1}{c|}{P-value}  	     		&\multicolumn{1}{c|}{Score} 	&  \\ \cline{2-9}
	& \multicolumn{1}{|l|}{code2seq}     				&{32.07} 		& {14.11}   			&\multicolumn{1}{l|}{<0.01} 	    		& {47.65}          	& {20.59}        			&\multicolumn{1}{l|}{<0.01}                		&\multicolumn{1}{l|}{16.71}  	&    \\
	& \multicolumn{1}{|l|}{attendgru}     				&{34.14} 		& {{\color{white}0}3.05}   	&\multicolumn{1}{l|}{<0.01} 	    		& {51.13}          	& {{\color{white}0}2.31}    	&\multicolumn{1}{l|}{<0.01}                		&\multicolumn{1}{l|}{19.07}  	&    \\
	& \multicolumn{1}{|l|}{transformer}     			&{33.97} 		& {{\color{white}0}3.87}   	&\multicolumn{1}{l|}{<0.01} 			& {51.16}         	& {{\color{white}0}2.06}     	&\multicolumn{1}{l|}{<0.01}                		&\multicolumn{1}{l|}{18.48}  	&    \\
	& \multicolumn{1}{|l|}{codegnngru}     			&{33.22} 		& {{\color{white}0}8.07}   	&\multicolumn{1}{l|}{<0.01} 	 		& {49.06}          	& {12.49}        			&\multicolumn{1}{l|}{{\color{white}0}0.02}	&\multicolumn{1}{l|}{18.03}  	&    \\
	& \multicolumn{1}{|l|}{HANcode}     				&{27.38} 		& {36.22}   			&\multicolumn{1}{l|}{<0.01} 	     		& {40.33}          	& {53.52}        			&\multicolumn{1}{l|}{<0.01}                		&\multicolumn{1}{l|}{14.07}  	&    \\
	& \multicolumn{1}{|l|}{SMNcode}     				&\textbf{34.68} 		& {-}   				&\multicolumn{1}{l|}{-} 				& {\textbf51.53}          	& {-}        				&\multicolumn{1}{l|}{-}                		&\multicolumn{1}{l|}{\textbf{19.47}}  	&    \\ \cline{2-9}
\end{tabular}

\begin{tabular}{llllllllllll}
 	\cline{2-9}
	& \multicolumn{1}{|p{3cm}|}{Python} 								& \multicolumn{3}{p{3cm}|}{METEOR}   							& \multicolumn{3}{p{3cm}|}{USE}                                        			&\multicolumn{1}{p{1cm}|}{BLEU}  	&  \\
	& \multicolumn{1}{|p{3cm}|}{\textbf{Models}}         	&{Score} 		& {T-test}   			&\multicolumn{1}{c|}{P-value}       		& {Score} 		& {T-test}        			&\multicolumn{1}{c|}{P-value}  	     		&\multicolumn{1}{c|}{Score} 	&  \\ \cline{2-9}
	& \multicolumn{1}{|l|}{code2seq}     				&{13.38} 		& {51.33}   			&\multicolumn{1}{l|}{<0.01} 	    		& {28.05}          	& {39.94}        			&\multicolumn{1}{l|}{<0.01}                		&\multicolumn{1}{l|}{{\color{white}0}5.02}  	&    \\
	& \multicolumn{1}{|l|}{attendgru}     				&{26.25} 		& {{\color{white}0}3.81}   	&\multicolumn{1}{l|}{<0.01} 	    		& {41.28}          	& {{\color{white}0}2.26}    	&\multicolumn{1}{l|}{{\color{white}0}0.01}  	&\multicolumn{1}{l|}{17.62}  	&    \\
	& \multicolumn{1}{|l|}{transformer}     			&\textbf{27.96} 		&{{\color{white}0}4.91}   	&\multicolumn{1}{l|}{<0.01} 			& \textbf{42.43}         	& {{\color{white}0}3.79}     	&\multicolumn{1}{l|}{<0.01}                		&\multicolumn{1}{l|}{19.70}  	&    \\
	& \multicolumn{1}{|l|}{codegnngru}     			&{25.85} 		& {10.01}   			&\multicolumn{1}{l|}{<0.01} 	 		& {39.15}          	& {{\color{white}0}9.21}    	&\multicolumn{1}{l|}{<0.01}			&\multicolumn{1}{l|}{19.12}  	&    \\
	& \multicolumn{1}{|l|}{HANcode}     				&{23.56} 		& {15.56}   			&\multicolumn{1}{l|}{<0.01} 	     		& {36.90}          	& {22.87}        			&\multicolumn{1}{l|}{<0.01}                		&\multicolumn{1}{l|}{16.16}  	&    \\
	& \multicolumn{1}{|l|}{SMNcode}     				&{27.01} 		& {-}   				&\multicolumn{1}{l|}{-} 				& {41.71}          	& {-}        				&\multicolumn{1}{l|}{-}                		&\multicolumn{1}{l|}{\textbf{19.81}}  	&    \\ \cline{2-9}

\end{tabular}

\label{tab:basic}
\vspace{-0.3cm}
\end{table*}

\vspace{-0.1cm}
\subsection{RQ2: Investigation of the improved summaries}
\label{sec:rq2}

To investigate the summaries improved by our encoder we 1) compare our non-ensembled encoder against basic models, and 2) discuss the features of the improved set.

We observe that our SMN encoder without ensemble with a non-statement based approach achieves higher USE, METEOR, and BLEU scores compared to all the basic models for Java dataset and all but one basic model for the Python dataset. We show the results of our evaluation in Table~\ref{tab:basic}. The top subtable reports the results for the Java dataset and bottom subtable reports the results for the Python dataset. The statistical t-tests are performed between our approach {\small \texttt{SMNcode}} and other models, so we only report those for the basic models and our baseline.

For the Java dataset we found {\small \texttt{SMNcode}} outperforms every other approach. The biggest difference we observe is against {\small \texttt{HANcode}}, another statement based approach, with a 7.30 METEOR, 11.20 USE, and 5.40 BLEU score improvement. We posit that {\small \texttt{HANcode}} relies on the pre-trained embedding mentioned in their paper or requires a much larger dataset to learn statement level information compared to {\small \texttt{SMNcode}}. We also observe that {\small \texttt{SMNcode}} achieves significantly higher score than every model with a P-value <0.02. This observation indicates that although we combine our statement based encoder with other approaches in order to take advantage of previous advancements in literature, our encoder can be perform just as well or even better than state-of-the-art independently. These results corroborate our initial results, that memory networks encode statement-based information better than HAN.

For the Python dataset we found that {\small \texttt{SMNcode}} outperforms every baseline except {\small \texttt{transformer}}. We observe that {\small \texttt{SMNcode}} achieves 3.45 METEOR, 4.81 USE, and  3.65 BLEU higher than {\small \texttt{HANcode}}. We also observe that {\small \texttt{transformer}} achieves  0.95 METEOR, 0.72 USE higher than our approach, but 0.11 BLEU lower than our approach. It appears that in contrast to the Java dataset, Python dataset benefits from token level self-attention more than statement-level self-attention overall. Recall from Section~\ref{sec:rq2} we noticed some unique features of the Python dataset. We posit that the small set of methods that are easier to summarize may benefit from copy-attention mechanism in the transformer models. Python dataset also has a larger number of challenging summaries that are harder to summarize because of lack of a generalizable structure which may further benefit from token-level self-attention. However, as we observed in RQ1,  {\small \texttt{transformer}} is further improved by addition of {\small \texttt{SMNcode}}. The ensemble {\small \texttt{transformer+SMNcode}} achieves 8\% higher scores compared to {\small \texttt{transformer}}. We suspect that for a subset of summaries, statement-level information is more important than token-level information. We test our hypothesis further by evaluating this improved subset in Table~\ref{tab:improvement}.

\begin{table}[h]
\centering
\vspace{-0.2cm}
\caption{METEOR scores for the improved set, where our approach SMNcode achieves higher scores when compared to the basic models and our baseline.}
\vspace{-0.2cm}
\begin{tabular}{lllllllllll} 
 	\cline{2-8}
	& \multicolumn{1}{|p{1.5cm}|}{}                                 	& \multicolumn{3}{c|}{Java}                                                                               & \multicolumn{3}{c|}{Python}                       &  \\
	& \multicolumn{1}{|p{1.5cm}|}{\textbf{Models}}  	& {Size} 		& {Base}    	&\multicolumn{1}{c|}{SMN}  	&{Size} 		& {Base} 		&\multicolumn{1}{c|}{SMN} &  \\ \cline{2-8}
	& \multicolumn{1}{|l|}{code2seq}     			& {47.20}          		& {25.21}        	&\multicolumn{1}{l|}{39.80}                	&{53.10} 		& {10.73} 		&\multicolumn{1}{l|}{41.97}  	&    \\
	& \multicolumn{1}{|l|}{attendgru}     			& {40.43}         		& {23.27}        	&\multicolumn{1}{l|}{36.69}    		&{38.19} 		& {17.99}   	&\multicolumn{1}{l|}{33.89}  	&    \\
	& \multicolumn{1}{|l|}{transformer}     		& {42.24}         		& {25.18}        	&\multicolumn{1}{l|}{38.87}                	&{36.28} 		& {16.09}   	&\multicolumn{1}{l|}{29.95}  	&    \\
	& \multicolumn{1}{|l|}{codegnngru}     		& {44.21}          		& {24.11}        	&\multicolumn{1}{l|}{37.99}		&{39.90} 		& {12.94}   	&\multicolumn{1}{l|}{28.74}  	&    \\
	& \multicolumn{1}{|l|}{HANcode}     			& {61.21}         		& {22.81}        	&\multicolumn{1}{l|}{39.78}                	&{44.66} 		& {13.15}   	&\multicolumn{1}{l|}{31.36}  	&    \\ \cline{2-8}
\end{tabular}
\vspace{-0.4cm}
\label{tab:improvement}
\end{table}

\begin{table*}[t]
	\centering
	\caption{Comparison of different configuration and statistical tests against best-performing configuration SMNcode}
\begin{tabular}{llllllllllll} 
 	\cline{2-9}
	& \multicolumn{1}{|p{3cm}|}{Java}                                 	& \multicolumn{3}{p{3cm}|}{METEOR}                                                                                  & \multicolumn{3}{p{3cm}|}{USE}                                        				&\multicolumn{1}{p{1cm}|}{BLEU}   	&  \\
	& \multicolumn{1}{|p{1.85cm}|}{\textbf{Models}}         	& {Score} 		& {T-test}        	&\multicolumn{1}{c|}{P-value}  	     			&{Score} 	& {T-test}   	&\multicolumn{1}{c|}{P-value}       		&\multicolumn{1}{c|}{Score} 	&  \\ \cline{2-9}
	& \multicolumn{1}{|l|}{SMNcode}     				& {34.68}         	& {-}        		&\multicolumn{1}{l|}{-}                			&{51.53} 	& {-}   		&\multicolumn{1}{l|}{-} 	    			&\multicolumn{1}{l|}{19.47}  	&    \\
	& \multicolumn{1}{|l|}{SMNcode-H1}     			& {34.39}         	& {2.66}    		&\multicolumn{1}{l|}{<0.01}              			&{50.85}	& {3.84}  		&\multicolumn{1}{l|}{<0.01}	    		&\multicolumn{1}{l|}{19.11} 	&    \\
	& \multicolumn{1}{|l|}{SMNcode-H2}     			& {34.62}          	& {0.37}       	&\multicolumn{1}{l|}{{\color{white}0}0.35}  		&{51.35}	& {1.04}  		&\multicolumn{1}{l|}{{\color{white}0}0.14}	&\multicolumn{1}{l|}{19.23} 	&    \\
	& \multicolumn{1}{|l|}{SMNcode-H4}     			& {34.45}          	& {1.25}       	&\multicolumn{1}{l|}{{\color{white}0}0.10}         	&{50.66}	& {4.91}  		&\multicolumn{1}{l|}{<0.01}	    		&\multicolumn{1}{l|}{19.42} 	&    \\
	& \multicolumn{1}{|l|}{SMNcode-H5}     			& {34.15}         	& {3.03}       	&\multicolumn{1}{l|}{<0.01}              			&{51.27}	& {1.49}  		&\multicolumn{1}{l|}{{\color{white}0}0.07}	&\multicolumn{1}{l|}{19.07} 	&    \\
	& \multicolumn{1}{|l|}{SMNcode-summary vector}     	& {34.06}         	& {3.57}       	&\multicolumn{1}{l|}{<0.01}              			&{50.28}	& {7.18}  		&\multicolumn{1}{l|}{<0.01}	    		&\multicolumn{1}{l|}{18.92} 	&    \\
	& \multicolumn{1}{|l|}{SMNcode-EOS Encoding}     	& {34.18}         	& {2.76}       	&\multicolumn{1}{l|}{<0.01}              			&{50.54}	& {5.60}  		&\multicolumn{1}{l|}{<0.01}	    		&\multicolumn{1}{l|}{19.01} 	&    \\ \cline{2-9}
\end{tabular}
\label{tab:smnrq3}
\vspace{-0.3cm}
\end{table*}

The way to interpret Table ~\ref{tab:improvement} is that for the first entry, there is a subset of  47.20\% of the methods in the Java dataset where our standalone encoder {\small \texttt{SMNcode}} performs better than {\small \texttt{code2seq}} where the average METEOR score for our approach is 39.80 compared to 25.21 by {\small \texttt{code2seq}}. We observe that compared to the {\small \texttt{transformer}} baseline there are 36.28 percent of the functions in the Python dataset where our encoder achieves higher METEOR,USE, and BLEU scores. This observation confirms our suspicion as for subset  {\small \texttt{SMNcode}} achieves almost 90\% higher METOR scores on average than {\small \texttt{transformer}}. This observation also means there is a comparable subset where {\small \texttt{transformer}} does better in order to get slightly higher METEOR and USE  score in Table~\ref{tab:basic}. The ensembles in RQ1 benefit from both of these subsets and hence achieve the highest score.

\vspace{-0.1cm}
\subsection{RQ3: Design Decisions and Configurations}

In Table~\ref{tab:smnrq3} we compare the different design decisions and configurations against our best performing approach {\small \texttt{SMNcode}}. For our first test, we observe that $h=3$ iterations provide achieves the best overall performance. We find that {\small \texttt{SMNcode-H2}} where $h=2$ performs almost as well as {\small \texttt{SMNcode}} where $h=3$. The difference between these could not be confirmed as statistically significant with a p-value greater than 0.05. While  {\small \texttt{SMNcode-H4}} is a close third, we see that {\small \texttt{SMNcode-H5}} gets the lowest scores The performance the performance peaks around $h=3$. We posit that {\small \texttt{SMNcode-H2}} attends to the two most important statements in the input and while a third statement adds some improvement, this selection of the most important facts leads of the improvements shown by our approach. Therefore, too many memory iterations decreases the quality of this selection. The size of the model and training time also increase with the number of iterations, this could make $h=2$ a good candidate for future work if there are resource constraints.

For the second test, we see that {\small \texttt{SMN-summary vector}}  achieves lower scores than our best performing configuration. We posit that because the summary vector already updates the GRU within the memory network because of RNN back-propagation, it does not add any new information for the memory network. In fact, the scores are even lower than the baseline {\small \texttt{attendgru}}. This suggests that summary so far may be acting as a distractor to the memory network. However, a fixed $Q$ matrix in {\small \texttt{SMNcode}} allows the model to learn which statements are more important using self-attention between statements within the memory network.

For the third test, we find that positional encoding achieves significantly higher scores for all three metrics compared to EOS encoding. We observe POS encoding in {\small \texttt{SMNcode}}  achieves 0.5 METEOR, 0.99 USE and 0.46 BLEU improvements over the EOS encoding. With p-values below 0.01 these improvements are statistically significant. We find that EOS encoding saw improvements in natural language application~\cite{kumar2016ask} , it links every statement with the representation of the previous statement. However, it does not encode the order in which the statement occurs in code. We posit that this explicit positional information better informs the memory network as source code is executed sequentially.

\vspace{-0.1cm}
\section{Discussion\slash Future Work}

This paper improves upon  state-of-the-art in four ways. First, a memory network outperforms very recent baseline work in encoding statement-level information for source code summarization. Second, statement based memory network improves a subset of summaries significantly, while other subsets may benefit from other approaches. Multiple approaches can be ensembled to take advantage of these orthogonal improvements. Third, memory network relies on learning which statements are more important to the summary among all statements. Each memory gives maximum attention to one statement. We posit that too many memories can decrease performance. Fourth, explicit encoding of the position of a statement is crucial to model performance.

Our approach outperforms recent baseline work in statement-based encoders on datasets of two different programming languages.  Our approach also improves several state of the art non-statement-based approaches. Additional intellectual merit of this paper is in the evaluation of the subsets of both datasets where different approaches excel. We explore the difference set for the purpose of intellectual curiosity. The results provide an insight into both the dataset and the advantages of different approaches. Future work could benefit from this way of evaluating source code summarization models. We see that different models improve niche subset of summaries and our approach improves every other model.

One possible avenue of future work is to apply memory networks to external context such as approaches described in Section~\ref{sub:codesummary}. Bansal~\textit{et al.}~\cite{bansal2021project} show that specific subroutines in various files in a project can help generate better summaries. A memory network could learn which of these subroutines is most important and could produce interesting results.

\vspace{-0.1cm}
\section{Conclusion}

This paper introduces a statement based approach to source code summarization using a memory network designed for the task. We provide a thorough evaluation on the potential of statement based approaches and memory networks to improve automated source code summarization. We show how our approach outperforms recent baselines using automated metrics suggested by the latest research literature. We investigated the subset of summaries that are most improved by our approach. We perform statistical tests for each of our results to test the significance of our results and reduce the risk of random differences in observation. We evaluated different configurations and design choices to arrive at the best performing configuration. We show how the best configuration outperform other possible design choices.

We conclude that both statement based approaches and memory networks can improve source code summarization in a way that is different from other approaches so far. We suggest multiple avenues for future work to explore how these techniques can be incorporated to improve the state-of-the-art.
\vspace{-0.25cm}
\section{Reproducibility}
\label{sec:repo}
To encourage reproducibility and future work by other researchers, we provide all our code, datasets, predictions, and instructions for replication using an online repository:

\emph{github.com/aakashba/smncode2022}

\section*{Acknowledgment}
This work is supported in part by NSF CCF-2100035 and CCF-2211428. Any opinions, findings, and conclusions expressed herein are the authors and do not necessarily reflect those of the sponsors.

\bibliographystyle{IEEEtran}
\bibliography{main}

\end{document}